%% file: main.tex
\documentclass[10pt,twocolumn,letterpaper]{article}

\usepackage{iccv}              

\input{preamble}

\definecolor{iccvblue}{rgb}{0.21,0.49,0.74}
\usepackage[pagebackref,breaklinks,colorlinks,allcolors=iccvblue]{hyperref}

\def\paperID{9486} 
\def\confName{ICCV}
\def\confYear{2025}

\title{Thermal Polarimetric Multi-view Stereo}

\author{Takahiro Kushida \quad Kenichiro Tanaka\\
Ritsumeikan University\\
2-150 Iwakura-cho, Ibaraki, Osaka, Japan\\
{\tt\small \{tkushida, ken-t\}@fc.ritsumei.ac.jp}
}

\begin{document}
\maketitle
\input{sec/abstract}
\input{sec/intro}

\input{sec/relatedwork}

\input{sec/physics}

\input{sec/proposed}
\input{sec/experiment}
\input{sec/conclusion}

\section*{Acknowledgment}
This work was supported by JSPS KAKENHI Grant Numbers
JP24K20802, 
JP22K18420, 
JP24K02965. 

{
    \small
    \bibliographystyle{ieeenat_fullname}
    \bibliography{main}
}

\end{document}

%% file: preamble.tex
\usepackage{siunitx}
\usepackage{bm}
\usepackage{colortbl}

\newcommand{\bbradiance}{B}
\newcommand{\wavelength}{\lambda}
\newcommand{\temperature}{T}
\newcommand{\planck}{h}
\newcommand{\speedoflight}{c}
\newcommand{\boltzmann}{k}
\newcommand{\emissivity}{\varepsilon}

\newcommand{\reflectance}{r}
\newcommand{\transmittance}{t}
\newcommand{\mullermatrix}{\bm{M}}
\newcommand{\transmittancemuller}{\bm{F}^{\mathrm{T}}}
\newcommand{\emissivitymuller}{\bm{\mathcal{E}}}
\newcommand{\emissionmuller}{\bm{E}}
\newcommand{\reflectancemuller}{\bm{F}^\mathrm{R}}

\newcommand{\zenithinsideobj}{\zenith'}
\newcommand{\refractiveindex}{\eta}
\newcommand{\rotatecoordinate}{\bm{C}}
\newcommand{\azimuth}{\varphi}
\newcommand{\stokesvec}{\bm{s}}
\newcommand{\stokes}{s}

\newcommand{\surfacenormal}{\bm{n}}
\newcommand{\rotation}{\bm{r}}
\newcommand{\rotationmat}{\bm{R}}
\newcommand{\transpose}{\mathsf{T}}
\newcommand{\tangentvec}{\bm{t}}

\newcommand{\surfacepoint}{\bm{x}}
\newcommand{\projectiontocamera}{\Pi}
\newcommand{\numviews}{N}
\newcommand{\loss}{\mathcal{L}}
\newcommand{\lossweight}{\lambda}
\newcommand{\numpixelsamples}{P}
\newcommand{\surfacepointspace}{\bm{X}}
\newcommand{\networkparam}{\bm{\theta}}
\newcommand{\visibility}{\Phi}
\newcommand{\crossentropy}{\Psi}
\newcommand{\sharpness}{\alpha}
\newcommand{\sigmoid}{\sigma}
\newcommand{\mask}{O}
\newcommand{\distmaskobj}{f^*}
\newcommand{\eikonal}{\mathbb{E}}
\newcommand{\zenith}{\theta}
\newcommand{\networkparamdims}{d}
\newcommand{\realspace}{\mathbb{R}}
\newcommand{\surface}{\mathcal{M}}
\newcommand{\mlp}{f}

\newcommand{\bbstokes}{\stokesvec_B}
\newcommand{\depolarizationmuller}{\bm{D}}

\definecolor{yesgreen}{rgb}{ .4, .9, .4}
\definecolor{moderateyellow}{rgb}{ .9, .9, .1}
\definecolor{nored}{rgb}{ .92, .5, .5}
\newcommand{\cg}{\cellcolor{yesgreen}}
\newcommand{\cy}{\cellcolor{moderateyellow}}
\newcommand{\crd}{\cellcolor{nored}}

\newcommand{\TODO}[1]{\textbf{\color{red}[TODO: #1]}}

\renewcommand{\TODO}[1]{}

%% file: sec/abstract.tex
\begin{abstract}
This paper introduces a novel method for detailed 3D shape reconstruction utilizing thermal polarization cues. Unlike state-of-the-art methods, the proposed approach is independent of illumination and material properties. In this paper, we formulate a general theory of polarization observation and show that long-wave infrared (LWIR) polarimetric imaging is free from the ambiguities that affect visible polarization analyses. Subsequently, we propose a method for recovering detailed 3D shapes using multi-view thermal polarimetric images. Experimental results demonstrate that our approach effectively reconstructs fine details in transparent, translucent, and heterogeneous objects, outperforming existing techniques.
\end{abstract}

%% file: sec/intro.tex
\section{Introduction}
3D shape reconstruction is a fundamental problem in computer vision, with applications spanning both academia and industry. 
Over the years, various 3D reconstruction methods using RGB cameras have been proposed, but many of these methods rely on strong assumptions about illumination and reflection models. For instance, multi-view stereo~\cite{schonbergerStructurefromMotionRevisited2016, schonbergerPixelwiseViewSelection2016, hartleyMultipleViewGeometry2003, szeliskiComputerVisionAlgorithms2022} depends on surface texture to find correspondences, while structured light~\cite{scharsteinHighaccuracyStereoDepth2003, gengStructuredlight3DSurface2011}, photometric stereo~\cite{durouComprehensiveIntroductionPhotometric2020,santoDeepPhotometricStereo2022,woodhamPhotometricMethodDetermining1980,hertzmannExamplebasedPhotometricStereo2005}, and shape from polarization~\cite{shiRecentProgressShape2020, miyazakiPolarizationbasedInverseRendering2003, wolffConstrainingObjectFeatures1991, miyazakiTransparentSurfaceModeling2004, atkinsonRecoverySurfaceOrientation2006, baDeepShapePolarization2020} require specific lighting conditions and assume opaque surfaces, limiting their applicability to a narrow range of materials.

Thermal imaging presents an attractive alternative for 3D reconstruction in challenging scenarios because it does not rely on environmental illumination; any object with heat emits long-wave infrared~(LWIR) light, effectively serving as its own light source.
Furthermore, most materials, except for those designed for thermal optics, are opaque in the LWIR spectrum, eliminating the need for many of the assumptions required by traditional reflection models.
Building on this concept, several methods, such as thermal multi-view stereo~\cite{zhangPhotometricCorrectionInfrared2023} and absorption-based depth estimation~\cite{nagaseShapeThermalRadiation2022,kushidaAffineTransformRepresentation2024,dorkengallastegiAbsorptionbasedHyperspectralThermal2024,gallastegiAbsorptionBasedPassiveRange2025} have been proposed. Although these methods demonstrate potential applicability, their results also imply the limitation inherent in model-based approaches, especially with regard to accuracy.

Photometric cues in the LWIR spectrum are useful for recovering the accurate shape, similar to the visible-light spectrum. 
For example, thermal photometric stereo~\cite{tanakaTimeResolvedFarInfrared2021} and shape from heat conduction~\cite{narayananShapeHeatConduction2024} have demonstrated detailed surface reconstruction. 
However, these methods require active heating and cooling of the object, making them both time-consuming and impractical.
In this paper, we aim to reconstruct fine 3D shapes in a steady state using polarimetric cues, a type of photometric cue that can be obtained without monitoring the heating or cooling processes.

\Cref{tab:method_comparison} presents a comparison of various 3D reconstruction approaches. Our approach is illumination-independent, material-independent, free from heating or cooling process, and highly accurate.
Our contributions are twofold. 
First, we establish a unified theoretical framework for LWIR polarization, showing that, unlike visible-light polarization, it is not affected by specular-diffuse ambiguities.
Second, we show that LWIR polarization serves as a powerful cue for detailed 3D shape reconstruction. Experimental results confirm that our method outperforms the existing approaches on heterogeneous materials.

%% file: sec/relatedwork.tex
\section{Related Work}

\begin{table*}
    \centering
    \footnotesize
        \begin{tabular}{|c|c|c|c|c|}
        \hline
         Method             & Illumination   & Material    & Accuracy &  Measurement   \\
         \hline
         Visible Multi-view Stereo~\cite{schonbergerStructurefromMotionRevisited2016, schonbergerPixelwiseViewSelection2016, hartleyMultipleViewGeometry2003, szeliskiComputerVisionAlgorithms2022}         & \crd Dependent    & \crd Textured only     & \cg High   &  \cg Easy  \\
         Visible SL / PS ~\cite{scharsteinHighaccuracyStereoDepth2003, gengStructuredlight3DSurface2011,durouComprehensiveIntroductionPhotometric2020,santoDeepPhotometricStereo2022,woodhamPhotometricMethodDetermining1980,hertzmannExamplebasedPhotometricStereo2005}              & \crd Dependent    & \crd Opaque only     & \cg High    &  \cg Easy  \\
         Visible Polarization~\cite{shiRecentProgressShape2020, miyazakiPolarizationbasedInverseRendering2003, wolffConstrainingObjectFeatures1991, miyazakiTransparentSurfaceModeling2004, atkinsonRecoverySurfaceOrientation2006, baDeepShapePolarization2020}   & \crd Dependent    & \crd Dependent      & \cg High  &  \cg Easy  \\
         Visible NeRFs~\cite{mildenhall2020nerf, yariv2020multiview}  & \crd Dependent    & \cg Wide range of materials     & \crd Low   &  \cg Easy  \\
         Thermal Multi-view Stereo~\cite{zhangPhotometricCorrectionInfrared2023}          & \cg Independent   & \crd Textured only & \cg High    & \cg Easy  \\
         Thermal SL~\cite{erdozain3dImagingThermal2020}                 & \cg Independent   & \crd Diffuse reflective only       & \cg High  & \cg Easy  \\
         Thermal PS~\cite{tanakaTimeResolvedFarInfrared2021}              & \cg Independent    & \cg Wide range of materials      & \cy Moderate  & \crd Heating \& cooling \\
         
         Thermal NeRFs~\cite{yeThermalNeRFNeuralRadiance2024,linThermalNeRFThermalRadiance2024}            & \cg Independent   & \cg Wide range of materials       & \crd Low  & \cg Easy   \\
         Heat Conduction~\cite{narayananShapeHeatConduction2024}         & \cg Independent   & \cg Wide range of materials       & \cy Moderate  &  \crd Heating  \\
         Thermal Polarization~(\textbf{Ours})          & \cg Independent   & \cg Wide range of materials & \cg High  &  \cg Easy   \\
         \hline
    \end{tabular}

    \caption{Comparison table of shape reconstruction methods. While other methods have pros and cons, our approach is independent of illumination and material's optical properties as well as accurate shape recovery is possible in steady state measurements.}
    \label{tab:method_comparison}
\end{table*}

\subsection{Shape from visible polarization}
Shape from Polarization (SfP), which recovers surface normals from the polarization state of reflected light, has a long-standing history in computer vision~\cite{shiRecentProgressShape2020, miyazakiPolarizationbasedInverseRendering2003, wolffConstrainingObjectFeatures1991, miyazakiTransparentSurfaceModeling2004, atkinsonRecoverySurfaceOrientation2006, baDeepShapePolarization2020}. 
Recently, the advent of snapshot polarization cameras has significantly boosted this area of research~\cite{rodriguezPola4AllSurveyPolarimetric2024}.

SfP approaches continue to tackle ambiguities in surface normals that arise from the mixture of specular and diffuse polarizations~\cite{shiRecentProgressShape2020}.
One effective way to resolve these ambiguities is to use multi-view polarization images, which provide a richer set of constraints on the surface normals and lead to more accurate and robust surface reconstructions~\cite{cuiPolarimetricMultiviewStereo2017}.
Another approach involves using polarimetric BRDF models, which can handle complex material properties~\cite{baekSimultaneousAcquisitionPolarimetric2018, zhaoPolarimetricMultiViewInverse2023, fukaoPolarimetricNormalStereo2021, hwangSparseEllipsometryPortable2022}
.
Recent approaches incorporate neural scene representations and differentiable rendering, enabling the joint estimation of shape, environment maps, and reflectance~\cite{liNeISFNeuralIncident2024, hanNeRSPNeural3D2024}.
Nevertheless, SfP in the visible light spectrum remains fundamentally sensitive to the illumination conditions, such as the assumption of uniform, unpolarized ambient light, and to variations in material's optical properties, including transparency, translucency, matte, or shiny finishes~\cite{shiRecentProgressShape2020}.

In this paper, we introduce the use of polarization in the LWIR spectrum. The LWIR light is emitted directly from the object and is independent of external illumination. Moreover, the polarization of LWIR emission is not mixed up with the diffuse and specular reflections or transmission effects. 
These properties make LWIR polarization advantageous for robust shape reconstruction.

\subsection{LWIR imaging and physics-based vision}
Beyond 3D reconstruction, thermal imaging is utilized for a variety of physics-based vision tasks. The analysis of heat conduction, for instance, has been leveraged for material classification by exploiting differences in thermal diffusivity and conductivity~\cite{saponaroMaterialClassificationThermal2015}.
Reflection analysis of LWIR light has facilitated non-line-of-sight (NLoS) imaging, expanding scene understanding by examining how thermal rays propagate around the corner~\cite{kagaThermalNonlineofsightImaging2019, maedaThermalNonLineofSightImaging2019}.
Joint analyses of both visible and LWIR light transport have further enriched scene understanding~\cite{ramanagopalTheoryJointLight2024}. 
We also advance the area of physics-based thermal vision through a shape reconstruction using LWIR polarization.

Thermal imaging is an effective approach for 3D shape reconstruction in some challenging scenarios. For example, it is effective for stealth observation as the thermal radiation does not rely on external illumination~\cite{nagaseShapeThermalRadiation2022, chenThermal3DGSPhysicsInduced3D2025, yeThermalNeRFNeuralRadiance2024, zhangPhotometricCorrectionInfrared2023}. Another example is that it is feasible to observe the shape of challenging materials in the visible light spectrum because many materials that appear transparent or translucent are opaque in the LWIR spectrum~\cite{tanakaTimeResolvedFarInfrared2021}.

Existing 3D reconstruction techniques using thermal cameras can be broadly categorized into geometric and photometric approaches. Geometric methods~\cite{mitaRobust3DPerception2019, zhangPhotometricCorrectionInfrared2023,erenScanningHeating3D2009} such as based on stereo/multi-view observations recover surface depth by matching corresponding points across thermal images, commonly assuming near-Lambertian emission or sufficient texture.
Photometric approaches infer depth, shape, or surface normal from intensity cues, which include thermal photometric stereo~\cite{tanakaTimeResolvedFarInfrared2021}, absorption-based method~\cite{nagaseShapeThermalRadiation2022,kushidaAffineTransformRepresentation2024,dorkengallastegiAbsorptionbasedHyperspectralThermal2024,gallastegiAbsorptionBasedPassiveRange2025}, and heat conduction-based method~\cite{narayananShapeHeatConduction2024}.
While they show strong possibilities of shape reconstruction using a thermal camera, 
their accuracy is generally limited.
There are hybrid approaches that combine visible-light and LWIR polarization~\cite{miyazakiTransparentSurfaceModeling2004}, where the LWIR polarization is used only as an auxiliary cue to resolve the specular–diffuse ambiguity in visible polarization rather than directly contributing to shape recovery.

Recently, neural rendering methods inspired by Neural Radiance Fields~(NeRF) have been explored in the thermal domain~\cite{yeThermalNeRFNeuralRadiance2024, linThermalNeRFThermalRadiance2024, chenThermal3DGSPhysicsInduced3D2025} as well. By jointly optimizing a volumetric scene representation and a rendering function to satisfy the rendered images and align observed images from multiple viewpoints, these thermal NeRF techniques recover both geometry and radiometric properties.
We are also inspired by this neural representation and model the object's surface by a neural implicit surface.

In this paper, we leverage the polarization state of thermal radiation to enrich the information available for 3D reconstruction. By incorporating the polarization cues, geometric rotation, and neural representation, our method enables more accurate shape estimation.

%% file: sec/physics.tex
\section{Polarimetric LWIR Observation}

In this section, we present a general theory of polarization observation and discuss the advantages of thermal polarization over visible polarization.

\subsection{Physics of Polarization}

The polarization state of light, including both visible and LWIR spectra, changes when it is reflected from or transmitted through an object's surface. The transport of polarized light can be described using the Stokes parameters $\stokesvec=[\stokes_0, \stokes_1, \stokes_2, \stokes_3]^\transpose$ and the Mueller matrix $\mullermatrix$ as 
\begin{equation}
    \stokesvec_o = \mullermatrix \stokesvec_i ,
\end{equation}
where $\stokesvec_i$ and $\stokesvec_o$ denote the Stokes parameters of the incoming and outgoing light, respectively, both defined in the surface coordinate system~\cite{collinComputationPolarizedSubsurface2014}. 

The observed scene is a combination of specular reflection, diffuse reflection, transmission, and emission. 
Consequently, the overall polarization state $\stokesvec$ can be expressed as
\begin{align}
    \stokesvec = \stokesvec_s + \stokesvec_d + \stokesvec_t + \stokesvec_e,
\end{align}
where $\stokesvec_s$, $\stokesvec_d$, $\stokesvec_t$, and $\stokesvec_e$ represent the Stokes vectors corresponding to the specular, diffuse, transmission, and emission components, respectively.

\begin{figure}[t]
    \centering
    \includegraphics[width=0.9\linewidth]{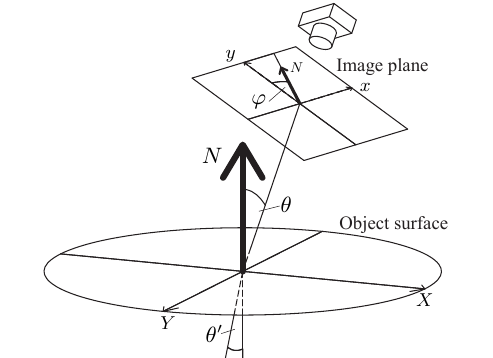}
    \caption{Coordinate systems and the zenith and projected azimuth angles of the surface normal. While Mueller calculus is performed in the object's surface coordinate system, the observed Stokes parameter and projected azimuth angle are represented in the image coordinate system.}
    \label{fig:model}
\end{figure}

\paragraph{Specular polarization} The change in the polarization state due to specular reflection~\cite{baekSimultaneousAcquisitionPolarimetric2018,ichikawaShapeSkyPolarimetric2021} is described by
\begin{align}
    \stokesvec_s = 
    \rotatecoordinate(\azimuth) \reflectancemuller(\zenith)  \stokesvec_i,
\end{align}
where $\stokesvec_i$ denotes the Stokes vector of the incident light in the object's surface coordinate system, 
$\stokesvec_s$ denotes the corresponding Stokes vector in the image coordinate system, 
$\rotatecoordinate(\azimuth)$ is a rotation matrix that transforms coordinates from the image to the object's surface, 
$\azimuth$ is the projected azimuth angle~(\ie the angle between the vertical axis of the image plane and the surface normal projected into the image plane), $\reflectancemuller(\zenith)$ is the Mueller matrix for the specular component, and $\zenith$ is the zenith angle of the reflection~(\ie the angle between the surface normal and the viewing direction).
The surface coordinate system, image coordinate system, zenith angle, and projected azimuth angle are illus in \cref{fig:model}.
The rotation matrix $\rotatecoordinate(\azimuth)$ is given by
\begin{equation}
    \rotatecoordinate(\azimuth) =
    \begin{bmatrix}
        1 & 0 & 0 & 0 \\
        0 & \cos2\azimuth & -\sin2\azimuth & 0 \\
        0 & \sin2\azimuth & \cos2\azimuth & 0 \\
        0 & 0 & 0 & 1 
    \end{bmatrix}.
\end{equation}
The Mueller matrix $\reflectancemuller(\zenith)$ is derived from Fresnel's law and expressed as
\begin{equation}
    \reflectancemuller(\zenith) = 
    \begin{bmatrix}
        \reflectance_{+} & \reflectance_{-} & 0 & 0 \\
        \reflectance_{-} & \reflectance_{+} & 0 & 0 \\
        0 & 0 & \reflectance_{\times} \cos\delta & \reflectance_{\times}\sin\delta \\
        0 & 0 & -\reflectance_{\times}\sin\delta & \reflectance_{\times} \cos\delta
    \end{bmatrix},
\end{equation}
where $\reflectance_{\pm}=\frac{\reflectance_s \pm \reflectance_p}{2}$, $\reflectance_{\times}=\sqrt{\reflectance_p\reflectance_s}$, $\zenithinsideobj$ denotes the refractive angle 
Here, $\cos\delta$ equals $-1$ when 
$\zenith$ is less than Brewster's angle and $1$ otherwise, with the sign of $\sin\delta$ reversed accordingly~\cite{baekSimultaneousAcquisitionPolarimetric2018}.
The Fresnel reflectances for s- and p-polarization, $\reflectance_s$ and $\reflectance_p$, are given by 
\begin{align}
    \label{eq:reflectances}
    \reflectance_s(\zenith) = \left( \frac{\cos\zenith - \refractiveindex\cos\zenithinsideobj}{\cos\zenith+\refractiveindex\cos\zenithinsideobj} \right)^2, 
    \reflectance_p(\zenith) = \left( \frac{\cos\zenithinsideobj - \refractiveindex\cos\zenith}{\cos\zenithinsideobj+\refractiveindex\cos\zenith} \right)^2.
\end{align}
According to Snell's law, the refractive angle is given as
\begin{equation}
    \zenithinsideobj = \sin^{-1}\left( \frac{1}{\refractiveindex} \sin \zenith \right),
\end{equation}
where $\refractiveindex$ represents the refractive index of the object.

\paragraph{Diffuse polarization} The change in the polarization state due to diffuse reflection or subsurface scattering~\cite{atkinsonRecoverySurfaceOrientation2006,ichikawaShapeSkyPolarimetric2021} is described by
\begin{align}
    \stokesvec_d = \rotatecoordinate(\azimuth) \transmittancemuller(\zenith)\int_{\Omega} \depolarizationmuller \transmittancemuller(\zenith)  \stokesvec_{i, \omega} d\omega,
\end{align}
where $\stokesvec_{i, \omega}$ denotes the Stokes vector of incident light coming from the direction $\omega$. The matrix $\depolarizationmuller$, which describes the depolarization, is defined as
\begin{equation}
    \depolarizationmuller = 
    \begin{bmatrix}
        \rho & 0 & 0 & 0 \\
        0 & 0 & 0 & 0 \\
        0 & 0 & 0 & 0 \\
        0 & 0 & 0 & 0
    \end{bmatrix},
\end{equation}
where $\rho$ denotes the proportion of the depolarization.
The Mueller matrix for transmittance
$\transmittancemuller(\zenith)$ is given by
\begin{equation}
    \transmittancemuller(\zenith) = 
    \begin{bmatrix}
        \transmittance_{+} & \transmittance_{-} & 0 & 0 \\
        \transmittance_{-} & \transmittance_{+} & 0 & 0 \\
        0 & 0 & \transmittance_{\times}\cos\delta  & \transmittance_{\times}\sin\delta \\
        0 & 0 & -\transmittance_{\times}\sin\delta & \transmittance_{\times} \cos\delta
    \end{bmatrix},
\end{equation}
where $\transmittance_{\pm}=\frac{\transmittance_s \pm \transmittance_p}{2}$, $\transmittance_{\times}=\sqrt{\transmittance_p\transmittance_s}$. 
$\transmittance_p$ and $\transmittance_s$ denote the Fresnel transmittance coefficients for s- and p- polarization, respectively, and are expressed as
\begin{equation}
    \transmittance_s(\zenith) = \left(\frac{2\cos\zenith}{\cos\zenith + \refractiveindex \cos\zenithinsideobj} \right)^2,
    \transmittance_p(\zenith) = \left(\frac{2\cos\zenith}{\cos\zenithinsideobj+\refractiveindex\cos\zenith}\right)^2.
\end{equation}

\paragraph{Transmission polarization} 
The polarization state also changes due to light transmission. According to the Fresnel's law, the transmission polarization is described by
\begin{align}
    \stokesvec_t = \rotatecoordinate(\azimuth) \transmittancemuller(\zenith) \stokesvec_i,
\end{align}
where $\stokesvec_i$ denotes the Stokes vector of the light coming from under the surface.

\paragraph{Emission polarization}  
The polarization state of emitted light is described by
\begin{align}
    \label{eq:general_emission_component}
    \stokesvec_e &=  \rotatecoordinate(\azimuth)\emissionmuller(\zenith)\stokesvec_c,
\end{align}
where $\stokesvec_c$ denotes the Stokes vector of the emitted light including thermal radiation as well as any chemical or electrical emissions and $\emissionmuller(\zenith)$ is the Mueller matrix corresponding to the emission component. Since there are many types of light emission, the elements of the emission Mueller matrix are not specified explicitly in general.

\subsection{LWIR Polarization Observation}
\paragraph{Polarization of visible light}
Since objects other than light sources do not emit light in the visible spectrum, the emission component is generally ignorable. Therefore, the observed polarization can be modeled as
\begin{align}
    \stokesvec = \stokesvec_d + \stokesvec_s + \stokesvec_t.
\end{align}
The magnitude of each component varies in complex ways depending on the object's material properties and the illumination environment. 
This complexity makes polarization analyses more challenging, leading existing visible polarization studies to rely on various assumptions to facilitate the analysis.

\paragraph{Polarization of LWIR light}
Most objects are known to be opaque in the LWIR spectrum; therefore, the transmission component can be safely ignored. 
Moreover, an interesting property of LWIR light polarization is that the relative contributions of each component can be controlled by the object's temperature. In a typical environments, where no objects are extremely hot in the surroundings~(\ie, where there are no significant LWIR light sources in the surroundings) the magnitude of reflection components is small compared to that of the emission. In such cases, the LWIR polarization observation can be modeled as
\begin{equation}
    \label{eq:lwir_stokes_observation}
    \stokesvec = \stokesvec_e.
\end{equation}

Since LWIR emission primarily arises from black body radiation, Eq.~\eqref{eq:general_emission_component} can be rewritten as
\begin{equation}
    \label{eq:emission_component_black_body_radiation}
    \stokesvec_e = \rotatecoordinate(\azimuth)\emissivitymuller(\zenith)\bbstokes(\temperature),
\end{equation}
where
$\emissivitymuller(\zenith)$ denotes the Mueller matrix for black body radiation, and $\bbstokes(\temperature)$ represents the Stokes vector of the radiation, which can be expressed as
\begin{equation}
    \label{eq:stokes-vec-blackbody}
    \bbstokes(\temperature) = 
    \begin{bmatrix}
    \bbradiance(\temperature) \\
    0 \\
    0 \\
    0 
    \end{bmatrix},
\end{equation}
where $\bbradiance$ denotes the black body radiation. According to Planck's law, $\bbradiance$ is given by
\begin{equation}
    \bbradiance(\temperature) = \frac{2\planck\speedoflight^2}{\wavelength^5} \frac{1}{e^{\frac{\planck\speedoflight}{\wavelength\boltzmann\temperature}}-1},
\end{equation}
where $c$ is the speed of light, $\boltzmann$ is the Boltzmann constant, $\planck$ is the Planck constant, $\temperature$ is the object's absolute temperature, and $\wavelength$ is the observation wavelength.
\Cref{eq:stokes-vec-blackbody} demonstrates that the object's temperature $T$ controls the power of emission.

The Mueller matrix of black body radiation $\emissivitymuller(\zenith)$~\cite{miyazakiDeterminingSurfaceOrientations2002,lemasterPassivePolarimetricImaging2014a} is given by 
\begin{equation}
    \emissivitymuller(\zenith) = 
    \begin{bmatrix}
        \emissivity_{+} & 0 & 0 & 0 \\
        \emissivity_{-} & 0 & 0 & 0 \\
        0 & 0 & 0 & 0 \\
        0 & 0 & 0 & 0 
    \end{bmatrix},
    \label{eq:black-body_muller}
\end{equation}
where $\emissivity_{\pm}=\frac{\emissivity_s \pm \emissivity_p}{2}$. $\emissivity_p$ and $\emissivity_s$ denote the emissivities for s- and p-polarization, respectively.
According to the energy conservation law, also known as Kirchhoff's law, the sum of reflectance, transmittance, and emissivity is equal to 1. Since the transmittance is negligible, as discussed above, we obtain
\begin{align}
    \emissivity_{s}(\zenith) &= 1 - \reflectance_s(\zenith) \\
    \emissivity_{p}(\zenith) &= 1 - \reflectance_p(\zenith).    
\end{align}

In summary, in the LWIR spectrum, the Stokes observation $\stokesvec$ can be analytically represented without ambiguity, thereby simplifying the analysis of the polarization state, which is a significant advantage for shape reconstruction.

\paragraph{Polarization to surface normal:}
LWIR polarization provides cues for an object's surface normal, which can be determined analytically.
The degree of linear polarization~(DoLP) is related to the zenith angle $\zenith$ as follows:
\begin{align}
    \text{DoLP} &= \frac{\sqrt{\stokes_1^2 + \stokes_2^2}}{\stokes_0} \\ 
    &= \frac{\left(\refractiveindex-\frac{1}{\refractiveindex}\right)^2\sin^2\zenith}
    {2 + 2\refractiveindex^2 - \left(\refractiveindex+\frac{1}{\refractiveindex}\right)\sin^2\zenith + 4\cos\zenith\sqrt{\refractiveindex^2-\sin^2\zenith}}.
\end{align}
The angle of linear polarization~(AoLP) is equivalent to the azimuth angle $\azimuth$ as follows:
\begin{equation}
    \text{AoLP} = \frac{1}{2}\tan^{-1}\left(\frac{\stokes_2}{\stokes_1}\right) = \azimuth.
\end{equation}
These correspondences enable the reconstruction of the surface normal from polarization observations.
Especially, AoLP serves as a robust cue because it directly reflects the azimuth angle of the surface normal and is independent of material properties, whereas DoLP is material-dependent due to the inclusion of the refractive index term.

\subsection{Examples of Visible vs. LWIR Polarization}

\begin{figure}[t]
    \centering
    \includegraphics[width=\linewidth]{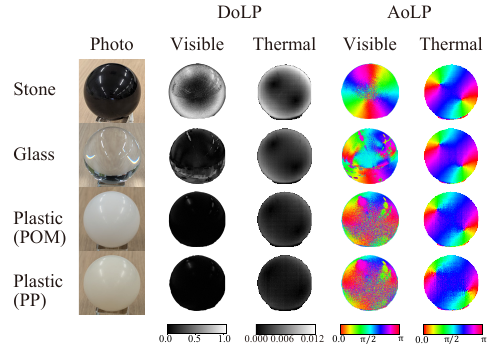}
    \caption{Visible and thermal polarization observations of spheres made from various materials under natural room lighting. While visible polarization is unstable due to variations in a material's optical properties, thermal polarimetric cues, especially AoLP images, remain consistent across different materials.
    }
    \label{fig:vis-vs-thermal}
\end{figure}

In the visible light spectrum, specular, diffuse, and transmission polarization components are often mixed~\cite{shiRecentProgressShape2020, baDeepShapePolarization2020, leiShapePolarizationComplex2022}, resulting in complicated observation.
\cref{fig:vis-vs-thermal} shows the polarimetric observations of spheres made from various materials; in the visible spectrum, unwanted polarization observations frequently appear. 
For example, because black stone exhibits strong specular polarization, the DoLP image is heavily affected by reflections from the surrounding environment. In the case of transparent glass sphere, the observation is notably contaminated by background textures. For translucent plastic materials, both DoLP and AoLP images are affected by surface texture and subsurface scattering. These unstable observations significantly complicate subsequent shape recovery.

In contrast, in the LWIR spectrum,  polarimetric observations remain consistent and independent of both material properties and environmental reflections, particularly in AoLP images. This consistency is a key advantage of thermal polarization cues, as it provides robustness against variations in materials and surrounding reflections.

%% file: sec/proposed.tex
\section{Thermal Polarimetric Multi-view Stereo}

Our goal is to recover the 3D shape using thermal polarimetric cues. As discussed in the previous section, the AoLP observation is invariant to material properties. To leverage this strong cue, we only use the AoLP images for material-independent 3D shape reconstruction.

Our approach employs multi-view thermal AoLP images to estimate the Signed Distance Function~(SDF), which is implicitly represented by a Multi-Layer Perceptron~(MLP) optimized via differentiable rendering. 
To effectively incorporate the surface normal information present in AoLP images, we adopt the concept of tangent space constraint~(TSC) proposed in a multi-view azimuth stereo~(MVAS)~\cite{caoMultiViewAzimuthStereo2023}.

\subsection*{Shape representation}
The SDF is implicitly represented by an MLP $\mlp(\surfacepoint;\networkparam): \realspace^3\times\realspace^\networkparamdims \rightarrow \realspace$, where $\surfacepoint\in\realspace$ denotes a point in space and $\networkparam\in\realspace^\networkparamdims$ are the learnable parameters of the MLP.
The zero-level set of the SDF defines the object's surface $\surface$ as~\cite{parkDeepSDFLearningContinuous2019}
\begin{equation}
    \surface(\networkparam) = \{\surfacepoint|\mlp(\surfacepoint;\networkparam)=0\}.
\end{equation}
The MLP is optimized within the framework of implicit differentiable renderer~(IDR)~\cite{yariv2020multiview}.

\subsection*{Loss function}
We adopt the loss function proposed in MVAS~\cite{caoMultiViewAzimuthStereo2023}, which is defined as,
\begin{equation}
    \loss = \loss_{\mathrm{TSC}} + \lossweight_1 \loss_{\mathrm{silhouette}} + \lossweight_2 \loss_{\mathrm{Eikonal}},
\end{equation}
where $\loss_\mathrm{TSC}$ is the tangent space consistency loss, $\loss_\mathrm{silhouette}$ is the silhouette loss and $\loss_\mathrm{Eikonal}$ is the Eikonal regularization. $\lossweight_1$ and $\lossweight_2$ are the weights assigned to the respective loss terms.

\paragraph{Tangent space consistency loss~\cite{caoMultiViewAzimuthStereo2023}}
The AoLP captured from different camera positions enforces a strong constraint on surface normal estimation. 
For a unit normal vector $\surfacenormal \subset \mathcal{S}^2 \in \mathbb{R}^3$ at a surface point $\surfacepoint \in \mathbb{R}^3$ and the projected azimuth angle $\azimuth$ at the corresponding pixel, the following relationship holds:
\begin{equation}
    \rotation_1^\transpose\surfacenormal\cos\azimuth - \rotation_2^\transpose\surfacenormal\sin\azimuth = 0,
\end{equation}
where $\rotationmat=[\rotation_1, \rotation_2, \rotation_3]^\transpose$ denotes the rotation matrix of the camera pose.
This equation can be rearranged as 
\begin{equation}
    \surfacenormal^\transpose 
    \underbrace{
    (\rotation_1 \cos \azimuth - \rotation_2 \sin \azimuth)
    }_{\tangentvec(\azimuth)}
    = 0,
    \label{eq:tsc}
\end{equation}
where $\tangentvec(\azimuth)$ is called a projected tangent vector and the above equation is called tangent space constraint. 
The TSC loss is the sum of the square error of the tangent space constraint and is defined as 
\begin{equation}
    \loss_{\mathrm{TSC}} = \frac{1}{\numpixelsamples} \sum_{\surfacepoint \in \surfacepointspace}
    \frac{\sum^{\numviews}_{i=1} \visibility_i \left(\surfacenormal^\transpose(\surfacepoint;\networkparam)\tangentvec_i(\surfacepoint)\right)^2}{\sum^\numviews_{i=1} \visibility_i}
\end{equation}
where $\numpixelsamples$ denotes the number of pixel samples, 
$\visibility_i$ is a binary indicator of the visibility of point $\surfacepoint$ from the $i$-th view, 
$\surfacenormal^\transpose(\surfacepoint;\networkparam)$ is the estimated surface normal at the point $\surfacepoint$ given network parameters $\networkparam$,
and $\tangentvec_i(\surfacepoint)$ is the tangent vector corresponding to the $i$-th view's pixel onto which the point $\surfacepoint$ is projected.

\paragraph{Silhouette loss~\cite{yariv2020multiview, caoMultiViewAzimuthStereo2023}}
The silhouette loss constrains the visual hull of the shape and is defined as
\begin{equation}
    \loss_{\mathrm{silhouette}} = \frac{1}{\sharpness\numpixelsamples} \sum_{\surfacepoint\in \tilde{\surfacepointspace}} \crossentropy\left(\mask(\projectiontocamera(\surfacepoint)), \sigmoid(\sharpness\distmaskobj)\right),
\end{equation}
where $\crossentropy$ denotes the cross entropy and $\mask$ is a binary mask indicating whether a pixel lies inside the object silhouette~(as provided by the input). 
$\numpixelsamples$ is the number of pixel samples,
$\projectiontocamera$ is the projection from the world coordinate to the image coordinate,
$\tilde{\surfacepointspace}$ represents the pixels with no ray-surface interactions or outside the silhouette,
$\distmaskobj$ represents the minimum SDF distance between the ray that does not have intersections and the object surface,
$\sigmoid$ is a sigmoid function, and $\sharpness$ controls the sharpness of the sigmoid function.

\paragraph{Eikonal regularization~\cite{groppImplicitGeometricRegularization2020}}
The Eikonal loss regularizes the gradient of SDF, enforcing the SDF to maintain its property of having a gradient close to 1:
\begin{equation}
    \loss_{\mathrm{Eikonal}} = \eikonal_{\surfacepoint}\left((||\surfacenormal(\surfacepoint;\networkparam)||_2 - 1)^2\right).
\end{equation}
$\eikonal_{\surfacepoint}(\cdot)$ denotes the expected value computed over sample points $\surfacepoint$ that are randomly sampled within the bounding box.

%% file: sec/experiment.tex
\section{Experiments}

\subsection{Setup}

\begin{figure}
    \centering
    \includegraphics[width=\linewidth]{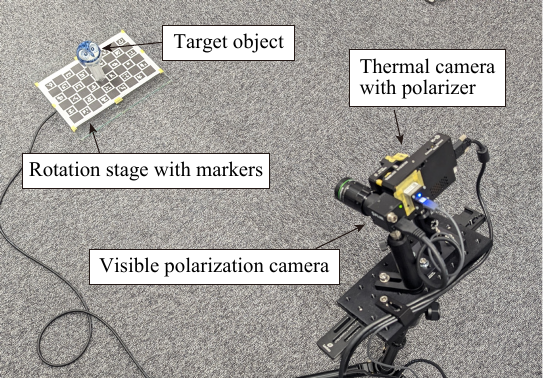}
    \caption{Camera system and setup. The system consists of a rotation stage to rotate the target and a thermal camera with a wire-grid polarizer. A visible polarization camera is used for camera pose estimation and for comparison with the visible MVAS.
    }
    \label{fig:setup}
\end{figure}

\begin{figure*}[t]
    \centering
    \includegraphics[width=\linewidth]{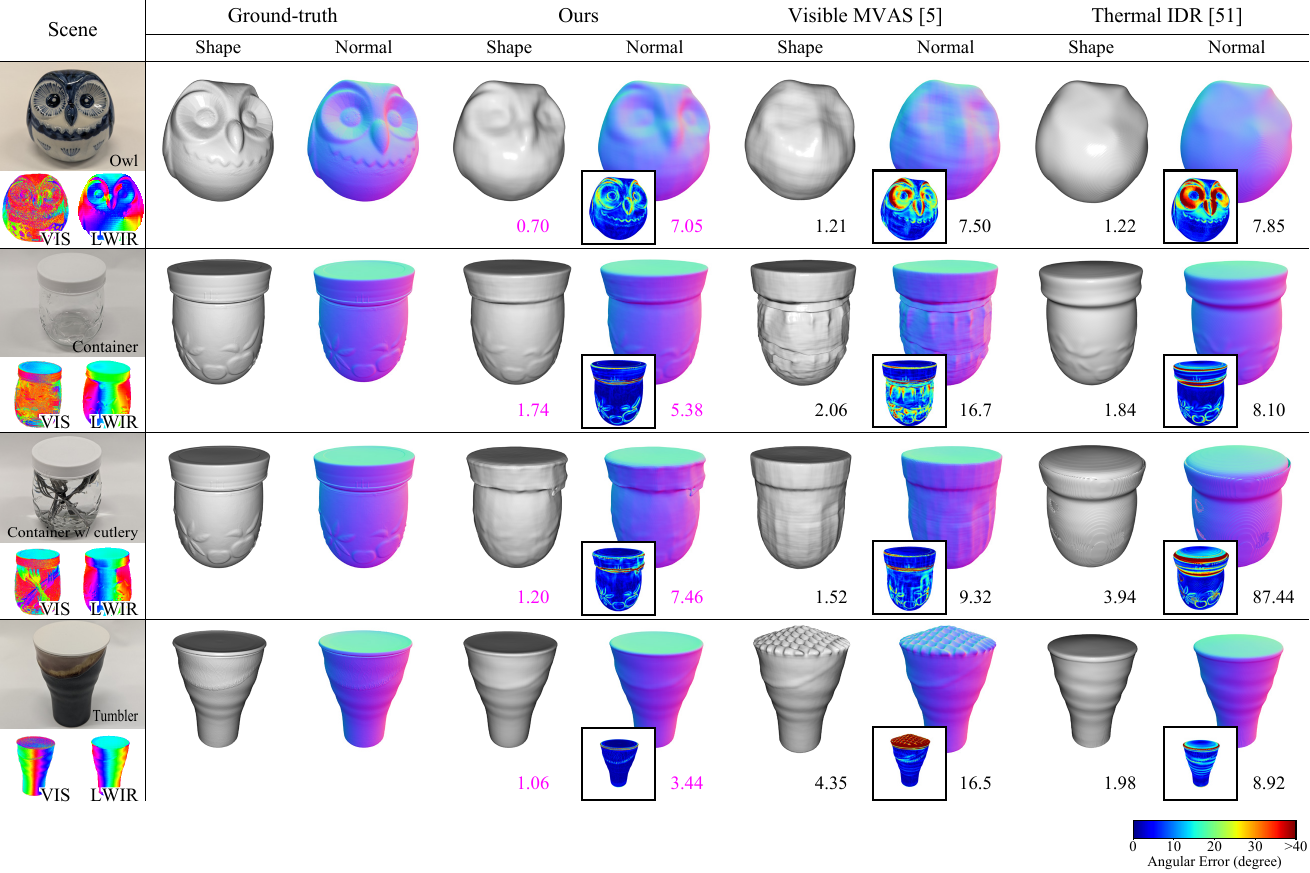}
    \caption{Experimental results. AoLP images in both visible and LWIR spectra are shown below the scene photograph. 
    Each method’s result includes the recovered shape, surface normals, and angular error maps.
    The Chamfer distance is displayed below the estimated shape and the mean angular error is displayed to the right of the angular error map. For both metrics, lower values indicate better performance. The smallest errors are shown in magenta. It demonstrates that the proposed method outperforms the other methods. 
    }
    \label{fig:results}
\end{figure*}

\paragraph{Implementation}
Following MVAS~\cite{caoMultiViewAzimuthStereo2023}, we use an 8-layer MLP with softplus activations, combined with positional encoding (dimension 10). The input 3D position and encoded features are skip-connected to the 4th layer of the network. The optimization is performed using the Adam optimizer with a learning rate of $0.001$. The weights for the loss terms are set to $\lambda_1 = 50$ (silhouette loss) and $\lambda_2 = 0.1$ (Eikonal loss). Training is conducted for 50 epochs with a batch size of 4096 pixels, and the silhouette loss weight is halved every 10 epochs.
Training takes approximately 2 hours on a single RTX A6000 GPU.

\paragraph{Camera system}
 \Cref{fig:setup} shows our camera system and experimental setup.
Our thermal polarization camera system consists of a thermal camera~(FLIR Boson 320) with \SI{15}{\milli\meter} lens and a wire-grid polarizer~(Thorlabs WP25M-IRC) mounted on a motorized rotation mount~(Thorlabs K10CR2).
The polarizer is installed between the image sensor and imaging lens to mitigate the narcissus reflections.
Thermal polarization images are captured by rotating the polarizer using the rotation mount. We captured four polarization images at 0, 45, 90, and 135 degrees.
A monochrome visible-light polarization camera~(FLIR Blackfly S BFS-U3-51S5P) is also placed adjacent to the thermal camera. The visible-light polarization camera is used both to capture images for a baseline method and to get the poses of both visible and thermal cameras.

\paragraph{Camera poses}

We employed a marker-based approach to estimate the camera poses. We put Aruco markers~\cite{garrido-juradoAutomaticGenerationDetection2014} on the rotation stage to estimate the pose of visible-light camera. Since printed markers are not visible in the thermal images, the poses of the thermal camera were obtained by transforming the camera poses of the visible-light camera. For stereo calibration between the visible-light and thermal cameras, we used a calibration target that can be observed by both cameras. The calibration board consists of a white-painted aluminum pegboard placed in front of a black surface heater. By heating the back panel, the circle grid pattern became visible to both cameras. The calibration parameters were estimated using Zhang's method~\cite{zhangFlexibleNewTechnique2000}.

\paragraph{Dataset}
We prepared a real-world dataset with ground-truth shapes.
We captured 7 objects with various shapes and materials. 
The target objects include transparent and low-reflective objects, which pose challenges for existing reconstruction methods.
The objects were placed on a rotating table and captured from about 20--30 views by rotating the table. 
To robustly capture thermal polarization, the objects were kept warm during the measurements. 
Masks for thermal images were generated by thresholding pixel values, while masks for visible images were generated by using Segment Anything Model 2~\cite{ravi2024sam2}. 
Ground-truth shapes were obtained using a structured light 3D scanner~(EinScan-SP V2). To facilitate the ground-truth measurement, the surfaces of the objects were painted white to make them opaque.

\paragraph{Baselines}

We compare our proposed method with two state-of-the-art methods: MVAS~\cite{caoMultiViewAzimuthStereo2023} based on visible-light polarization cues and 
IDR~\cite{yariv2020multiview}, which is one of the neural implicit surface reconstruction methods, with thermal image input.
For MVAS with visible-light polarization images, there exists $\pm\frac{\pi}{2}$ ambiguity in surface normals due to the mixture of specular and diffuse polarization. Therefore, TSC loss is modified to account for the $\pm\frac{\pi}{2}$ ambiguity, as originally proposed in MVAS. Since our method does not exhibit this ambiguity, the modified loss function is applied only to visible light observation.

\paragraph{Evaluation metrics}
We use two metrics to evaluate both the surface points and surface normals. Chamfer distance is the metric for evaluating the geometric error, which calculates the point cloud distance between the estimate and the ground-truth. The mean angular error is the metric for surface normal, which calculates the angular error of normal maps. For both metrics, a smaller value indicates a better result.
Prior to computing these metrics, the estimated and the ground-truth meshes are aligned using ICP algorithm~\cite{beslMethodRegistration3D1992}.

\subsection{Results}
\begin{figure}
    \centering
    \includegraphics[width=\linewidth]{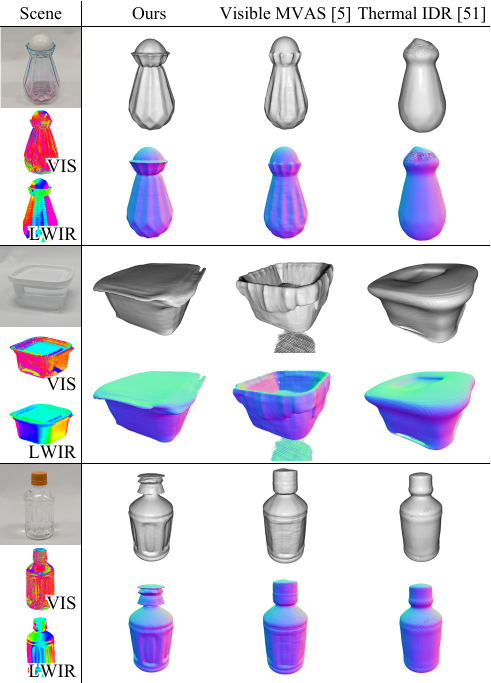}
    \caption{Qualitative results for other objects. AoLP images are shown below the scene photos. For each method, the upper image is the estimated shape and the lower is the corresponding normal map. It is shown that our method successfully reconstructs fine details of the object, particularly the concave parts in the transparent vase and bottle, while other methods fail to reconstruct such details. 
    }
    \label{fig:qualitative_results}
\end{figure}

\Cref{fig:results} shows the results. The target objects are an owl made of ceramic with a clear coating, a glass container with and without interiors, and a black tumbler, some of which are challenging objects for the existing approaches, especially a container with cutlery. In AoLP observations, while visible light is affected by a mixture of reflections and transmission components, thermal observations remain consistent and align with the azimuth angle of the surface normal. This consistency leads to a clear difference in detailed 3D reconstruction results between visible polarization and thermal polarization.

In the owl's result, while the detailed shape such as the eyes and the beak is missing in both visible MVAS and thermal IDR, they are successfully recovered in our result. Similarly, the relief pattern on the glass container is clearly visible in our results.
For a black tumbler with a white lid shown in the fourth row,
the AoLP observation on the lid in the visible-light spectrum is too noisy.
In such a case, the surface normal of the lid is not reconstructible, resulting in a wavy surface. In contrast, our method recovers a reasonable shape due to the presence of a unique AoLP on the lid in LWIR polarization. Overall, visible MVAS tends to reconstruct with wavy artifacts as AoLP is noisy, and thermal IDR tends to recover swollen shapes as it does not consider the surface normal. 

To quantitatively compare the results, the estimated shapes are evaluated using Chamfer distance, and the surface normals are evaluated by mean angular error. Our method outperforms the other methods in both Chamfer distance and the mean angular error for all objects.

\Cref{fig:qualitative_results} shows additional qualitative results. Our results are visually better than the other methods. The concave details of the vase and bottle are especially notable, as they are precisely reconstructed in our method while other results appear blurred. In addition, the surface of the container's lid is only reconstructed by our method while there is a big hole in the results of the others.
Some artifacts are seen in thin parts in our result, particularly in the neck of the bottle. This artifact comes from the low spatial resolution of the thermal camera, which can be reduced by using a higher-resolution camera or increasing camera views. 

%% file: sec/conclusion.tex
\section{Conclusion}
In this paper, we propose a 3D shape reconstruction using multi-view thermal polarimetric observations. 
We theoretically show that thermal polarimetric cues, especially AoLP images, are independent of material properties and illumination environment.
Based on this property, we demonstrate a detailed 3D shape recovery using multi-view thermal AoLP images and show that our approach is applicable to heterogeneous objects including transparent and low-reflective objects.

Although the proposed method performs well, some challenges remain in expanding the applicable scenes.
Our approach assumes the sufficient power of the emission component compared to the reflection component. This assumption is not valid when the hot objects are placed near the target object or when the target's temperature is lower than the environment.
Moreover, our method is less effective on metallic surfaces or rough surfaces, where the polarization signal can become unstable or noisy.
In such cases, decomposing the emission and reflection components~\cite{tanakaTimeResolvedFarInfrared2021} before applying the shape recovery or integrating additional cues may be a good option. 
Improving the observation system is also an important factor since our capturing system needs some time to rotate the polarization filter. A single-shot polarized LWIR camera could accelerate the thermal polarization analysis in the near future, as in the visible-light spectrum.